\documentclass[11pt]{article}

\usepackage[preprint]{acl}

\usepackage{times}
\usepackage{latexsym}
\usepackage{amsmath}
\usepackage{booktabs}
\usepackage{multirow}
\usepackage{graphicx}
\usepackage{amsfonts}

\usepackage[T1]{fontenc}

\usepackage[utf8]{inputenc}

\usepackage{microtype}

\usepackage{inconsolata}

\usepackage{graphicx}

\title{Where To Look? : Causal Tracing of Vision Encoders in VLM
}

\author{
 \textbf{Naren Kumar S},
 \textbf{Tirth Bhatt},
 \textbf{Mayank Singh}
\\
 LINGO Research Group, Indian Institute of Technology Gandhinagar, India
\\
 \small{
   \textbf{Correspondence:} \href{mailto:lingo@iitgn.ac.in}{lingo@iitgn.ac.in}
 }
}

\begin{document}
\maketitle
\begin{abstract}
Vision-language models can describe an image with remarkable accuracy, yet a more fundamental question remains unanswered: \textbf{what visual information actually drives their answers?} In this work, we investigate this question through causal tracing, and we observe that highly causal vision tokens often lie outside the target region. Extending the analysis to larger vision-language models reveals a similar pattern across models and corruption settings, suggesting that strong multimodal performance does not necessarily imply spatially localized causal representations. We further investigate: \textbf{can these models preserve visual structure when appearance cues are removed?} and find that visual cues are exploited to understand visual structures. Together, our experiments \textbf{expose a gap between seeing, using, and reasoning over visual structure}, and provide a causal framework for studying how visual information is transformed, preserved, and ultimately used by modern vision-language models
\end{abstract}

\section{Introduction}
Every year, frontier vision-language models (VLMs), both open-source and proprietary, achieve increasingly strong performance on benchmarks that evaluate challenging visual perception and reasoning capabilities. Yet, the emergence of new benchmarks often reveals substantial performance degradation, raising questions about whether these models genuinely acquire robust visual representations or instead exploit dataset-specific shortcuts. This discrepancy highlights that understanding how visual and linguistic inputs interact remains a challenge. Raising the question of whether these models rely on visual information encoded in the image or on shortcuts to answer questions. 

Recent studies in mechanistic interpretability have introduced causal tracing as a method to identify which internal component is responsible for a specific output. While these methods were primarily developed for language models, they are being heavily adopted to understand the behavior of VLMs. In this paper, we extend causal tracing to examine the behavior of the vision encoder in vision-language modeling, aiming to determine whether information about the image is lost throughout the process or not. We adapt the activation patching technique to the vision encoder and introduce a framework that links causal contribution $\Gamma$ with spatial alignment, measured via Intersection-over-Union (IoU) between token patches and ground-truth object regions.

\section{Related Work}

Using mechanistic interpretability, we investigate the internal computations underlying model behavior, with causal intervention methods enabling the identification of representations that directly influence model outputs. Causal tracing, initially developed for language models to localize information flow, has been extended to vision-language models, showing that intermediate visual representations can be causally implicated in multimodal predictions~\cite{causaltracingblip}. In parallel, visual grounding studies examine whether model representations or attention align with task-relevant image regions, with recent work showing that attention and spatial relevance can diverge and exploring explicit grounding methods for VLMs~\cite{visualgrounding, objectlocalization}. Rather than studying causal influence and spatial grounding separately, we directly examine their relationship by testing whether causally influential visual tokens align with regions required to answer a query.

We further investigate structural visual reasoning through a controlled string-tracing task, motivated by evidence that VLMs struggle to follow connected visual structures~\cite{visualgrounding}. By keeping string geometry and connectivity fixed while manipulating their appearance, the task isolates whether models preserve structural information beyond appearance-based cues. This complements the causal analysis by shifting from \emph{where} influential information is located to \emph{what visual structure} is retained by the vision encoder and made available for downstream reasoning. Together, these experiments probe whether multimodal predictions rely on spatially grounded and structurally faithful representations or instead on distributed, appearance driven cues.

\section{Methodology}
\label{sec:methodology}
In our experiment, we adapt causal tracing to the vision encoder to measure the causal influence of intermediate visual representations on the final image-text similarity score.

Given an image $I$ and a text query $T$, we can computes the image-text similarity as

\begin{equation}
s(I,T) = \langle f_{\text{img}}(I), f_{\text{text}}(T) \rangle .
\end{equation}

We use three forward passes to measure the causal influence of an intermediate activation:

\begin{itemize}
\item \textbf{Clean run:} We provide the original image $I$ to the model and record the resulting similarity score $s_{\text{clean}}$.

\item \textbf{Corrupted run:} We provide a corrupted version of the image, obtained by adding Gaussian noise, and record the resulting similarity score $s_{\text{corr}}$.

\item \textbf{Patched run:} We provide the corrupted image and replace a selected intermediate activation with the corresponding activation recorded during the clean run. We then record the resulting similarity score $s_{\text{patch}}$.

\end{itemize}

For a layer-token pair $(l,t)$, we define the causal contribution as

\begin{equation}
\Gamma_{l,t} =
\frac{s_{\text{patch}} - s_{\text{corr}}}
{s_{\text{clean}} - s_{\text{corr}}}.
\end{equation}

This quantity measures the fraction of the corrupted-to-clean similarity gap that is recovered by restoring the activation at layer $l$ and token $t$. A value of $\Gamma_{l,t}$ close to $1$ indicates that the activation has a strong causal effect on the similarity score. A value close to $0$ indicates that restoring the activation has little effect on the similarity score. Values outside the range $[0,1]$ can occur when activation patching produces over-recovery or a change that exceeds the clean-corrupted difference.

\subsection{Spatial Grounding Alignment}

Current state-of-the-art~(SOTA) vision-language models achieve strong performance on image understanding and visual grounding tasks. However, task performance alone does not establish whether the model uses the spatial region relevant to the input query. We therefore evaluate whether causally important vision tokens correspond to the spatial regions associated with the target object.

We map each vision token to its corresponding image patch. In a Vision Transformer (ViT)~\cite{vit}, each patch token represents a fixed spatial region of the input image. Let $P_t$ denote the image region associated with token $t$, and let $B$ denote the ground-truth bounding box associated with the referring expression. We measure the spatial correspondence between token $t$ and the target region using Intersection over Union (IoU):

\begin{equation}
\text{IoU}(P_t, B) = \frac{|P_t \cap B|}{|P_t \cup B|}
\end{equation}

We then compare the causal contribution of each token with its spatial overlap with the target region. Specifically, we compute the correlation between the causal contribution $\Gamma_{l,t}$ and $\text{IoU}(P_t,B)$:

\begin{equation}
\text{Corr}(\Gamma,\text{IoU}).
\end{equation}

A high positive correlation indicates that tokens with high causal contribution tend to correspond to spatially relevant image regions. A low correlation indicates that causal importance is weakly associated with the target region. This analysis therefore provides a quantitative measure of the relationship between causal importance and spatial grounding.

\subsection{Layer-wise Grounding Analysis}

To determine how spatial grounding develops across the vision encoder, we compute the correlation independently at each layer. For layer $l$, we define

\begin{equation}
\text{Corr}_l = \text{Corr}(\Gamma_{l,t}, \text{IoU}_t)
\end{equation}

This analysis enables us to identify the layers at which causal importance becomes aligned with the spatial region relevant to the input query. A systematic increase in $\text{Corr}_l$ across layers would indicate that spatial grounding develops progressively in the vision encoder. Alternatively, high correlation at intermediate or early layers would indicate that spatial information becomes causally relevant before the final layers.

Although current SOTA vision-language models use different vision encoders, Vision Transformers remain a common architectural component. Therefore, the proposed analysis provides a useful basis for studying spatial grounding in ViT-based vision encoders. However, the results should not be assumed to generalize directly to models that use substantially different visual architectures.

\section{Results}

We first investigate whether causally important visual representations are spatially aligned with the regions relevant to a given query. We investigate all VLMs from 2023 to the present. Beginning with CLIP~\cite{clip}, which provides a controlled setting for studying the relationship between image-text similarity and individual vision tokens. We then extend the analysis to larger vision-language models (VLMs) to determine whether the observations from CLIP persist in more capable multimodal models.

Our experiments are organized in three stages. First, we establish the causal-spatial relationship in CLIP. Second, we evaluate whether the same behavior is observed in larger VLMs and under different image corruption procedures. Third, we propose to extend the evaluation across model generations and visual encoder architectures, including both ViT-based and SigLIP-based encoders. For experiments with multiple random seeds, we report the mean and standard deviation across runs.

\subsection{Does Causally Important Token Strongly Align with Spatially Relevant Area}

We first evaluate CLIP~\cite{clip} using the causal tracing procedure described in Section~\ref{sec:methodology}. For each image-query pair, we compute the causal contribution $\Gamma_{l,t}$ of individual layer-token pairs and compare it with the spatial overlap between the corresponding image patch and the ground-truth bounding box.

Across 100 samples, we observe a weak relationship between causal importance and spatial alignment:

\begin{equation}
\mathrm{Corr}(\Gamma,\mathrm{IoU}) = 0.062.
\end{equation}

The low correlation indicates that tokens that have a strong causal effect on the image-text similarity score are not necessarily located within the spatial region associated with the target object. Conversely, tokens that overlap with the target object do not necessarily have a strong causal effect on the model output.

This observation is also visible at the individual token level. For example, in one representative example, a highly causal token achieves $\Gamma=0.718$, while its corresponding spatial overlap with the ground-truth bounding box is only $\mathrm{IoU}=0.049$. Several other highly ranked tokens also have zero overlap with the annotated target region.

\begin{figure}[t]
    \centering
    \includegraphics[width=1\linewidth]{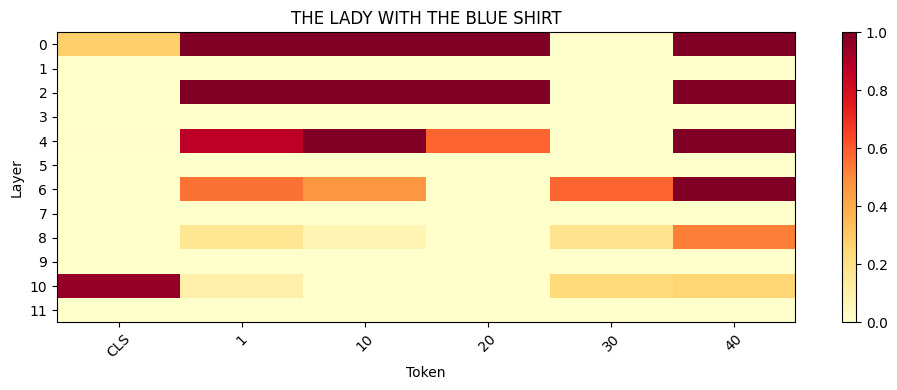}
    \caption{Causal contribution ($\Gamma$) across layers and vision tokens for the query ``the lady with the blue shirt''. The highly causal tokens are not necessarily located within the ground-truth target region.}
    \label{fig:heatmap}
\end{figure}

These results provide the first observation of our study that \textbf{\emph{causal importance does not necessarily imply spatial information is encoded}}. A token can strongly influence the model output even when the corresponding image patch has little overlap with the object in question.

\subsection{Layer-wise Behavior in CLIP}

We next examine whether the relationship between causal importance and spatial alignment changes across the layers of the vision encoder. For each layer $l$, we compute

\begin{equation}
\mathrm{Corr}_l =
\mathrm{Corr}(\Gamma_{l,t},\mathrm{IoU}_t).
\end{equation}

The correlation remains low across the vision encoder. We observe a small increase in the later layers, but the correlation remains weak even at its highest values.

\begin{figure}[t]
    \centering
    \includegraphics[width=1\linewidth]{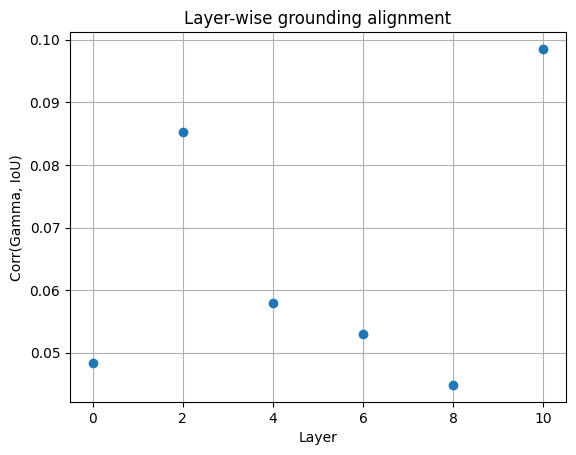}
    \caption{Layer-wise correlation between causal contribution and spatial alignment in the vision encoder.}
    \label{fig:layerwise}
\end{figure}

This result suggests that strong spatial grounding does not clearly emerge at a particular layer of the CLIP vision encoder. Instead, causal information appears to be distributed across visual representations that are not tightly localized to the target region.

The CLIP experiment motivates a broader question:

\begin{quote}
\textit{Does the weak relationship between causal importance and spatial alignment, found in larger and more capable vision-language models?}
\end{quote}

\begin{table*}[t]
\centering
\small
\setlength{\tabcolsep}{4pt}
\renewcommand{\arraystretch}{1.15}

\resizebox{\textwidth}{!}{%
\begin{tabular}{ll|ccccc}
\hline
\textbf{Architecture} &
\textbf{Corruption} &
\textbf{Corr. $\Gamma$--IoU} &
\textbf{Max $\Gamma$ Top-1} &
\textbf{Spatial Heuristic Top-1} &
\textbf{Intersection Top-$k$} &
\textbf{Jaccard Top-$k$} \\
\hline

\multirow{2}{*}{\textbf{DeepSeek-VL }}
& DDPM
& $0.0531 \pm 0.0334$
& $0.3865 \pm 0.1604$
& $0.0590 \pm 0.0490$
& $1.1737 \pm 0.0464$
& $0.0676 \pm 0.0030$ \\

& Blur
& $0.0520 \pm 0.0041$
& $0.4409 \pm 0.0058$
& $0.0362 \pm 0.0000$
& $1.1712 \pm 0.0474$
& $0.0678 \pm 0.0032$ \\
\hline

\multirow{2}{*}{\textbf{Qwen-2.5}}
& DDPM
& $0.0912 \pm 0.0422$
& $0.5365 \pm 0.3822$
& $0.0812 \pm 0.0509$
& $0.5948 \pm 0.0799$
& $0.0326 \pm 0.0045$ \\

& Blur
& $0.0294 \pm 0.0100$
& $0.8034 \pm 0.0868$
& $0.1806 \pm 0.0651$
& $0.5670 \pm 0.0486$
& $0.0311 \pm 0.0029$ \\
\hline

\multirow{2}{*}{\textbf{LLaVA-NeXT}}
& DDPM
& $0.0419 \pm 0.0166$
& $0.3167 \pm 0.1110$
& $0.0216 \pm 0.0118$
& $0.2135 \pm 0.0328$
& $0.0114 \pm 0.0018$ \\

& Blur
& $-0.0217 \pm 0.0052$
& $0.5065 \pm 0.0220$
& $-0.1152 \pm 0.0003$
& $0.1825 \pm 0.0157$
& $0.0098 \pm 0.0008$ \\
\hline

\multirow{2}{*}{\textbf{LLaVA-1.5}}
& DDPM
& $0.0509 \pm 0.0329$
& $0.9451 \pm 0.3710$
& $0.1357 \pm 0.0562$
& $2.0176 \pm 0.0417$
& $0.1244 \pm 0.0028$ \\

& Blur
& $0.0460 \pm 0.0001$
& $0.6367 \pm 0.0028$
& $0.0457 \pm 0.0021$
& $1.8140 \pm 0.0026$
& $0.1113 \pm 0.0001$ \\
\hline

\multirow{2}{*}{\textbf{InternVL}}
& DDPM
& $0.0397 \pm 0.0224$
& $0.3908 \pm 0.2166$
& $0.0826 \pm 0.1825$
& $0.0753 \pm 0.0153$
& $0.0041 \pm 0.0008$ \\

& Blur
& $0.0121 \pm 0.0033$
& $0.3187 \pm 0.0095$
& $0.0183 \pm 0.0000$
& $0.0727 \pm 0.0096$
& $0.0040 \pm 0.0005$ \\
\hline

\multirow{2}{*}{\textbf{SmolVLM}}
& DDPM
& $0.0011 \pm 0.0000$
& $0.1726 \pm 0.0000$
& $0.0803 \pm 0.0000$
& $0.0215 \pm 0.0000$
& $-0.0011 \pm 0.0000$ \\

& Blur
& $-0.0019 \pm 0.0000$
& $-0.4906 \pm 0.0000$
& $-0.0682 \pm 0.0000$
& $0.0289 \pm 0.0000$
& $0.0015 \pm 0.0000$ \\

\hline
\end{tabular}%
}

\caption{Causal importance and spatial alignment metrics across different VLM architectures under DDPM and blur corruptions. Values are reported as mean $\pm$ standard deviation over the evaluated runs.}
\label{tab:causal_spatial_results}
\end{table*}

\subsection{Extending the Analysis to Larger Vision-Language Models}

To investigate this question, we apply the same causal tracing procedure to larger VLMs. We evaluate DeepSeek-VL~\cite{deepseekvl}, Qwen-2.5~\cite{qwen2.5}, LLaVA-NeXT~\cite{llavanext}, LLaVA~\cite{llava1.5}, InternVL-2~\cite{internvl, internvl2}, and SmolVLM~\cite{smolvlm} using the same token-level causal intervention and spatial alignment procedure. The results are summarized in Table~\ref{tab:causal_spatial_results}. Overall, the evaluated VLMs exhibit the same qualitative behavior as CLIP. The correlation between causal contribution and spatial alignment remains low across models.

For DDPM corruption, the overall correlation is $0.0531 \pm 0.0334$ for DeepSeek-VL, $0.0912 \pm 0.0422$ for Qwen, $0.0419 \pm 0.0166$ for LLaVA-NeXT, $0.0509 \pm 0.0329$ for LLaVA, and $0.0397 \pm 0.0224$ for InternVL. These values remain close to the weak-correlation regime observed in CLIP.

The same trend is observed with blur corruption. DeepSeek-VL obtains $0.0520 \pm 0.0041$, Qwen-2.5 obtains $0.0294 \pm 0.0100$, LLaVA-NeXT obtains $-0.0217 \pm 0.0052$, LLaVA obtains $0.0460 \pm 0.0001$, and InternVL obtains $0.0121 \pm 0.0033$.

The results therefore provide preliminary evidence that the CLIP observation is not restricted to a contrastive image-text model. A similar lack of strong causal-spatial alignment is observed in larger VLMs.

\subsection{Causal Importance and Spatial Grounding Remain Distinct}

Although the overall correlations are low, several models contain tokens with substantial causal contributions. This is particularly clear for the maximum causal contribution among the top-ranked tokens. For example, under DDPM corruption, Qwen-2.5 obtains a mean maximum causal contribution of $0.5365 \pm 0.3822$, while DeepSeek-VL obtains $0.3865 \pm 0.1604$. Under blur corruption, these values increase to $0.8034 \pm 0.0868$ and $0.4409 \pm 0.0058$, respectively. These results are important because they show that the low correlation is not due to the absence of causally important tokens. Instead, the models contain tokens that strongly affect the output, but these tokens are not consistently aligned with the target object's spatial location.

Therefore, our analysis separates two properties of visual representations and proves that \textbf{\emph{a visual token can be causally important without being spatially localized to the region required by the query}}.

\subsection{Robustness to the Corruption Procedure}

The CLIP experiment was initially performed using image corruption to create a contrast between clean and corrupted model states. To determine whether our observations depend on a particular corruption mechanism, we extend the analysis to two corruption procedures: DDPM-based corruption and blur.

The overall causal-spatial correlation remains low for most models under both corruption settings. For example, DeepSeek-VL shows correlations of $0.0531 \pm 0.0334$ and $0.0520 \pm 0.0041$ under DDPM and blur, respectively. Qwen-2.5shows $0.0912 \pm 0.0422$ under DDPM and $0.0294 \pm 0.0100$ under blur. LLaVA shows a similar pattern, with $0.0509 \pm 0.0329$ under DDPM and $0.0460 \pm 0.0001$ under blur.

Some model-dependent variation is observed. LLaVA-NeXT, for example, changes from $0.0419 \pm 0.0166$ under DDPM corruption to $-0.0217 \pm 0.0052$ under blur corruption. However, the magnitude remains small in both cases. These results indicate that the weak causal-spatial relationship is not limited to a single corruption procedure. At the same time, the variation between corruption types shows that the intervention procedure can affect the measured causal contribution. We therefore treat corruption robustness as an important component of the evaluation rather than assuming that a single corruption mechanism is sufficient.

\subsection{Stability Across Random Seeds}

Causal tracing involves image corruption and intervention. A single run can therefore produce a noisy estimate of causal contribution. We address this issue by repeating the experiments with different random seeds. For DeepSeek-VL, Qwen-2.5, LLaVA-NeXT, and LLaVA, we perform 10 runs for each corruption setting. InternVL is evaluated using 6 runs in the current experiment. We report the mean and standard deviation to quantify the variation across runs. The low overall correlations remain consistent across the repeated runs for the evaluated models. This provides preliminary evidence that the observed mismatch between causal importance and spatial alignment is not caused by a single random seed. SmolVLM currently has only one run in the reported experiment. We therefore do not use this result to make a strong statement about statistical stability. Additional runs are required before drawing conclusions for this model.

\subsection{From CLIP to Current Vision-Language Models}

The progression from CLIP to larger VLMs provides an important observation. The weak causal-spatial relationship initially identified in CLIP is also observed in the larger VLMs evaluated in our current experiments. This raises a broader question: \textbf{\emph{Does increasing model capability lead to stronger causal spatial grounding?}}

Modern VLMs have substantially improved in image understanding, visual question answering, reasoning, and instruction following. However, improved task performance does not necessarily imply that the internal visual representations have become more spatially localized. We therefore plan to perform a longitudinal evaluation of models released across multiple generations (refer to Appendix~\ref{sec:appendix}), beginning with early vision-language models and extending to present state-of-the-art~(SOTA) models. This evaluation will allow us to examine whether the causal-spatial relationship changes as VLMs become larger and more capable.

Our main hypothesis is:

\begin{quote}
\textit{Increase in model capability doesn't imply the model is well grounded and the representations embed the complete spatial information}
\end{quote}

In other words, we investigate whether improvements in multimodal performance are accompanied by improvements in causal spatial grounding. If the correlation remains weak across model generations, this would suggest that stronger VLM performance emerges without a corresponding increase in localized causal representations.

\subsection{Across Vision Encoder Architectures}

The current experiments primarily focus on VLMs with ViT-based vision encoders. This provides a suitable starting point because ViT tokens correspond naturally to spatial image patches. Therefore, each token can be directly associated with an image region and evaluated using IoU. However, current VLMs use more than one family of vision encoders. In particular, SigLIP-based~\cite{siglip} encoders are also widely used in recent multimodal models. We therefore plan to extend the analysis from ViT-based encoders to SigLIP-based encoders. This extension is important because it allows us to distinguish between two possible explanations for the observed behavior.

First, the weak causal-spatial alignment may be a general property of modern vision-language representations. If similar results are obtained with both ViT-based and SigLIP-based encoders, this would support this interpretation.

Second, the behavior may depend on the visual encoder architecture. If the causal-spatial relationship differs systematically between ViT and SigLIP encoders, this would indicate that the architecture contributes to the formation of spatially grounded representations.

The planned evaluation will therefore cover two dimensions: model generation (the year and family it belongs to) and the backbone of the vision encoder architecture.  We will evaluate models from earlier generations through current SOTA VLMs and compare ViT-based and SigLIP-based visual encoders using the same causal tracing framework.

\subsection{Overall Findings}

The experiments provide a progressive set of observations.
\begin{itemize}
    \item First, our CLIP experiment shows a weak relationship between causal importance and spatial alignment, with an overall correlation of $0.062$ across 100 samples.
    \item Second, the same qualitative behavior is observed when the analysis is extended to larger VLMs. Across the evaluated models, the overall causal-spatial correlations remain low under both DDPM and blur corruption.
    \item Third, the presence of highly causal tokens does not imply that these tokens correspond to the target object. Some tokens can have a strong effect on the model output while having low or zero spatial overlap with the ground-truth region.
    \item Fourth, the observation remains present across multiple random seeds and corruption procedures, although the magnitude varies across models.

\end{itemize}

Together, these results suggest that strong vision-language performance does not necessarily require strongly localized causal visual representations. The models may instead rely on distributed visual information that is not restricted to the spatial region associated with the queried object. Our current findings, therefore, motivate a broader investigation across model generations and visual encoder architectures. In the next stage of the study, we will evaluate VLMs from early models through current SOTA models and extend the analysis from ViT-based encoders to SigLIP-based encoders. This will allow us to determine whether the observed causal-spatial mismatch is a general property of modern VLMs or specific model architectures and training paradigms.

\section{Conclusion and Future Work}
In this work, we investigate whether visual representations that are causally important to vision-language models are also spatially aligned with the regions relevant to a given query. The findings suggest that strong vision-language performance does not necessarily require causally important visual representations to be spatially localized to the queried object, and that causal importance and spatial grounding represent distinct properties of multimodal representations. We further observe variation across models and corruption procedures, motivating a broader evaluation across multiple model generations and visual encoder architectures. As future work, we plan to study models from early vision-language models through current SOTA models and extend the analysis from ViT-based encoders to SigLIP-based encoders. This will allow us to investigate whether causal spatial grounding improves as model capabilities increase and whether the observed behavior is a general property of modern vision-language models or depends on the underlying vision encoder architecture.

\bibliography{custom}

\appendix

\section{Structural Information Tracing with String Puzzles}
\label{sec:appendix}

The previous experiments show that causal importance in the vision encoder does not consistently correspond to the spatial region required by the query. This raises a further question: \textbf{\emph{what type of visual information is preserved by the vision encoder and made available to the language model?}} In particular, spatial localization alone does not establish whether the model preserves relationships between distant image regions. To investigate this question, we introduce a controlled visual reasoning task based on string tracing.

\subsection{String Puzzle Task}

\begin{figure}[t]
    \centering
    \includegraphics[width=1\linewidth]{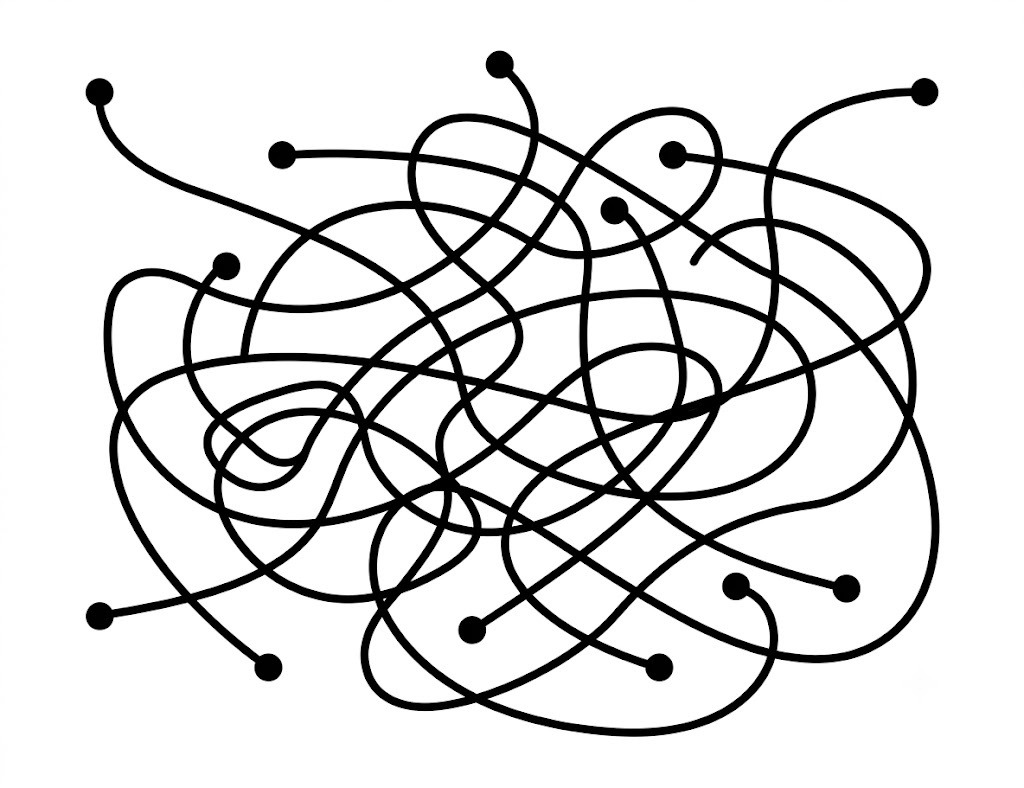}
    \caption{String-tracing puzzle in which all strings have the same color. The model must follow the continuous structure of the queried string to identify its endpoint.}
    \label{fig:string_same}
\end{figure}

We design a synthetic string-tracing task in which multiple visually overlapping strings connect pairs of endpoints. Given one endpoint, the model is asked to identify the endpoint at the other end of the same string. An example of the task is shown in Figure~\ref{fig:string_same}.

The task requires the model to follow a continuous visual structure through multiple crossings and determine the endpoint connected to the queried starting point. Unlike standard visual question answering tasks, the semantic content of the image is intentionally minimal. The primary information required to solve the task is the connectivity structure of the strings. We generate all puzzles using a custom Python-based generation procedure. The generator controls the number of strings, their trajectories, endpoint locations, and visual appearance. This allows us to construct paired examples with the same underlying string geometry while changing only the visual properties of the strings.

\subsection{Color-Consistent and Color-Distinct Settings}

We evaluate two versions of the string-tracing task. In the first setting, all strings have the same color. Therefore, the model cannot use color identity to distinguish one string from another. The only reliable cue for solving the task is the string's continuous geometric structure. In the second setting, each string is assigned a different color. The underlying geometry and endpoint connectivity remain unchanged. However, color provides an additional feature for distinguishing between strings.

This controlled design is important because the two conditions contain the same structural information. The primary difference is the availability of a low-level appearance cue. Therefore, the performance difference between the two conditions provides evidence of the type of information the model uses.

\begin{figure}[t]
    \centering
    \includegraphics[width=1\linewidth]{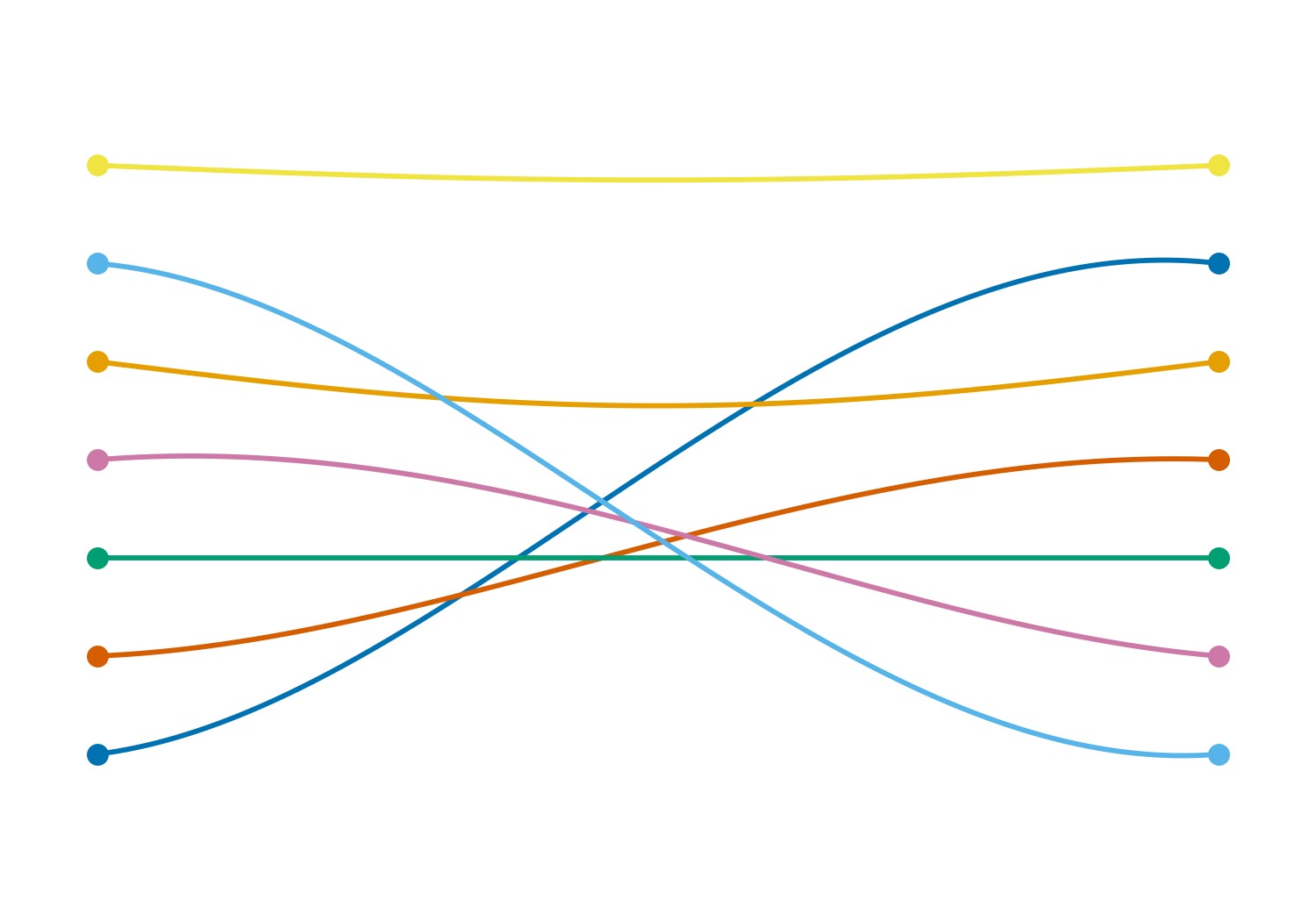}
    \caption{String-tracing puzzle with color-distinct strings. The underlying string geometry is unchanged, but color provides an additional cue for identifying the correct string.}
    \label{fig:string_color}
\end{figure}

\subsection{Evaluation}

For each puzzle, we provide the model with one endpoint and ask it to identify the endpoint connected to that point. We measure endpoint accuracy as the percentage of puzzles for which the model identifies the correct endpoint.

Let $y_i$ denote the correct endpoint for puzzle $i$ and $\hat{y}_i$ denote the model prediction. We define endpoint accuracy as

\begin{equation}
\mathrm{Accuracy}
=
\frac{1}{N}
\sum_{i=1}^{N}
\mathbb{I}(\hat{y}_i=y_i),
\end{equation}

where $N$ is the number of evaluated puzzles.

We compare performance between the same-color and color-distinct conditions. Since the underlying string topology is kept constant, a large performance difference between the two conditions indicates that color provides information that substantially facilitates the tracing task.

\begin{figure}[t]
    \centering
    \includegraphics[width=1\linewidth]{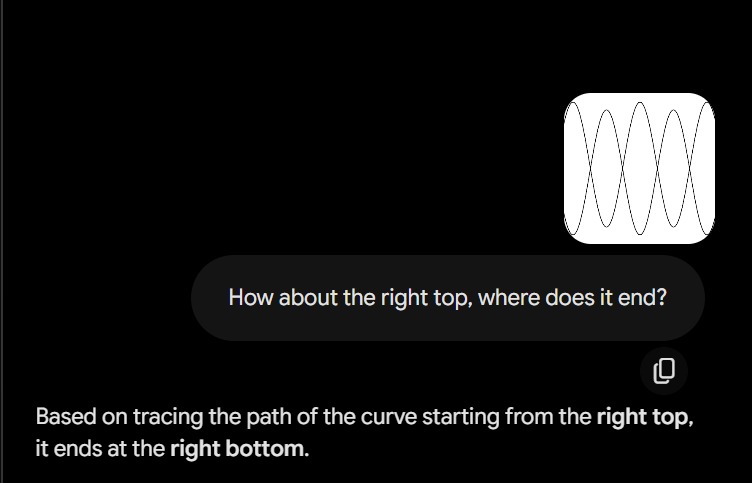}
    \caption{String-tracing puzzle in which all strings have the same color. The model is not able to distinguish properly. Even when only two interleaved strings are there}
    \label{fig:model_string_same}
\end{figure}

\begin{figure}[t]
    \centering
    \includegraphics[width=1\linewidth]{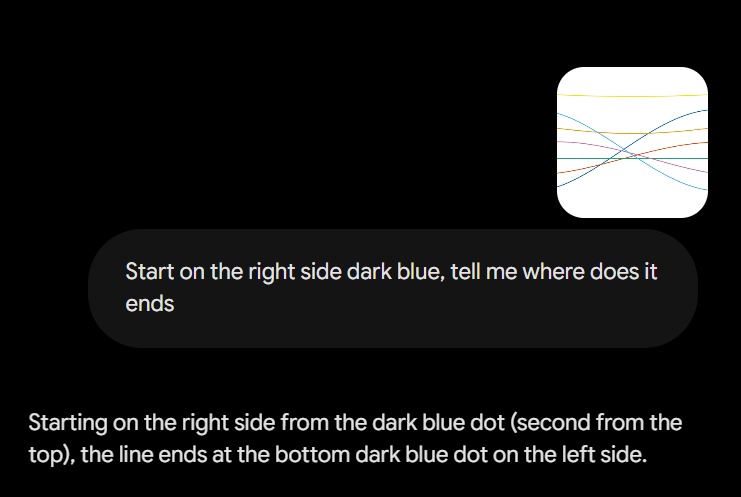}
    \caption{String-tracing puzzle with color-distinct strings. The model uses color as a cue and performs better.}
    \label{fig:model_string_color}
\end{figure}

\subsection{Initial Observations}

Our initial experiments, in which we infer proprietary models from a single instance of the dataset, we observe a substantial difference between the two conditions. When all strings have the same color, the model has difficulty identifying the correct endpoint. In contrast, when each string is assigned a distinct color, the model is able to trace the queried string and identify its corresponding endpoint more reliably. This difference is illustrated in Figure~\ref{fig:model_string_same} and Figure~\ref{fig:model_string_color}, the model used to evaluate is Gemini-3.1 Pro~\cite{gemini3.1pro} . In the color-distinct example, the model correctly follows the queried string from its starting endpoint to the corresponding endpoint. In the same-color setting, the model frequently fails to preserve the string's identity when multiple strings intersect or overlap.

Importantly, the underlying string structure is unchanged between the two conditions. Therefore, the difference in performance cannot be attributed to a change in the strings' actual connectivity. Instead, the result indicates that the availability of color information substantially affects the model's ability to solve the task. These observations suggest that the model may not reliably preserve the long-range structural identity of a string when strings have similar local visual appearance. When each string has a distinct color, that color serves as a persistent local feature that helps the model associate different parts of the same string across the image.

\subsection{Structural Information and Vision Encoder Representations}

The string-puzzle experiment provides a more controlled setting for studying the limitations suggested by our causal grounding experiments. In the previous experiments, the target object and its surrounding context could contain many semantic and visual cues. In contrast, the string puzzle contains minimal semantic information and requires the model to preserve a specific visual relationship across distant image regions.

The observed performance gap between the same-color and color-distinct conditions suggests that the vision encoder may not preserve structural connectivity in a sufficiently explicit form. \emph{When color differs between strings, the encoder can use differences in pixel values as a discriminative feature}. These features provide a relatively simple mechanism for maintaining the identity of a string across the image. In the same-color condition, this cue is removed. The model must instead infer connectivity from the geometry of the lines and their crossings. The reduced performance suggests that this structural information may be weakened or fragmented during visual encoding. We therefore hypothesize that \textbf{\emph{the vision encoder transforms the continuous string structure into visual tokens without fully preserving the long-range connectivity required for the task. As a result, the language model may receive local visual features that are individually informative but insufficient to reconstruct the complete structure of the string.}} This hypothesis is consistent with, but not established by, the accuracy results alone. To distinguish between information loss in the vision encoder and reasoning limitations in the language model, we perform additional analysis of intermediate visual representations and attention patterns.

\subsection{Attention Analysis}

To investigate how the model processes the string structure, we analyze the attention maps produced by the vision-language model. We examine whether visual attention follows the queried string as it moves through the image and whether attention is disrupted at string crossings.

For a query endpoint $e_s$, we examine the attention assigned to image tokens corresponding to the same string and compare it with attention assigned to visually similar but structurally unrelated strings. This analysis allows us to investigate two possible failure modes.
\begin{itemize}
    \item First, the vision encoder may fail to preserve a string's identity across distant image regions. In this case, attention would be expected to become distributed across multiple strings, particularly near crossing points. 
    \item Second, the visual representation may preserve sufficient information, but the language model may fail to use this information to perform the required multi-step tracing.
\end{itemize}
 In this case, the relevant visual information may be present in the encoded representations even when the final prediction is incorrect. We therefore use attention analysis as a complementary analysis rather than treating attention alone as evidence of causal reasoning.

\subsection{From Visual Encoding to Language-Model Reasoning}

The string puzzle also allows us to study the interaction between visual encoding and language-model reasoning. The task requires a sequence of operations: identify the queried endpoint, follow the corresponding string through the image, maintain its identity at crossings, and select the final endpoint. When the strings have different colors, the visual encoder provides a strong appearance-based identity signal. The language model can then use this signal to associate separated visual tokens with the same string. In contrast, when all strings have the same color, the model must rely primarily on geometric continuity.

This leads to a central question in our analysis:

\begin{quote}
\textit{When the visual encoder does not provide an explicit structural representation, does the language model reconstruct the missing structure through reasoning, or does it rely on local visual cues and guess the answer?}
\end{quote}

We investigate this question by comparing model predictions, visual attention patterns, and intermediate representations across the two conditions.

\subsection{Extension Across Model Generations}

Similar to our causal tracing experiments, we plan to evaluate the string-puzzle task across multiple generations of vision-language models. We will begin with early large-scale VLMs and extend the evaluation to current SOTA models. This evaluation will determine whether improved multimodal capabilities are accompanied by improved preservation of structural visual information. In particular, we will compare the performance gap between the same-color and color-distinct conditions:

\begin{equation}
\Delta_{\mathrm{color}}
=
\mathrm{Accuracy}_{\mathrm{distinct}}
-
\mathrm{Accuracy}_{\mathrm{same}}.
\end{equation}

A large value of $\Delta_{\mathrm{color}}$ indicates strong dependence on color-based cues, whereas a smaller value indicates that the model can solve the task using structural information even when color does not distinguish the strings. This provides a model-independent measure of the dependence on appearance cues.

\end{document}